# LSTM-RASA Based Agri Farm Assistant for Farmers

Narayana Darapaneni[1], Selvakumar Raj[2], Raghul V[3], Venkatesh Sivaraman[4], Sunil Mohan[5] and Anwesh Reddy Paduri[6]

[1] Northwestern University/Great Learning, Evanston, US
[2-6] Great Learning, Bangalore, India

anwesh@greatlearning.in

**Abstract.** The application of Deep Learning and Natural Language based ChatBots are growing rapidly in recent years. They are used in many fields like customer support, reservation system and as personal assistant. The Enterprises are using such ChatBots to serve their customers in a better and efficient manner. Even after such technological advancement, the expert advice doesn't reach the farmers on timely manner. The farmers are still largely dependent on their peer's knowledge in solving the problems they face in their field. These technologies have not been effectively used to give the required information to farmers on timely manner. This project aims to implement a closed domain ChatBot for the field of Agriculture – "Farmer's Assistant". Farmers can have conversation with the Chatbot and get the expert advice in their field. Farmer's Assistant is based on RASA Open Source Framework. The Chatbot identifies the intent and entity from user utterances and retrieve the remedy from the database and share it with the user. We tested the Bot with existing data and it showed promising results.

1. **Introduction**

Agriculture is one of the backbones of India. The employment percentage of Agricultural sector was around 77% and its contribution to GDP was 34% in 1983. Over the years, these numbers have fallen and they employ only 40% of Indian work force now and their share towards GDP has also declined. Many factors contribute to such drop in the importance of Agriculture. One of the main reasons is falling Return of Investment in the field of Agriculture. The ROI can be improved if Farmer's have sufficient access to expert advice on how to protect their crops, where can they get their seeds, what crops to farm in which season etc. Indian Government introduced Kisan Call center to provide such information to the farmers. The call centers operate in 22 languages, where experts are hired to clear the doubts of Farmers. The call center operates in 2 levels. Level 1 attends the call, gets the basic information from user and tries to provide the required information. The request will be transferred to Level 2 – Subject Matter Expert, with required information, when Level 1 couldn't clear the doubt. The SME has 72 hour window to revert back with the details requested. The KCC receives around 50 to 60 lakh calls every year. The call center resources cannot be expanded indefinitely when the demand grows. If the questions cannot be answered by Level 1, then the customer might have to wait for 72 hours to get the required information. Moreover, while hiring the experts the Government identifies the people who are knowledgeable in the crops prevalent to that Region. All these shortcomings can be overcome by building a ChatBot which possesses the knowledge of all the Agricultural experts in the region. Unlike call center, the ChatBot can cater to the growing demand or surge in calls. The user need not wait for 72 hour window for the expert to get back to them.

To experiment this, we build a ChatBot based on Rasa Framework. The ChatBot was trained on categories like plan protection and nutrient management on hand picked crops like Paddy, Coconut, tomato, banana and sugarcane.

## 2. Literature Review

### 2.1 Chatbot History

The discussion about Chatbot cannot happen without the mention of Alan Turing, who is widely considered as Father of Artificial Intelligence. He is the one who proposed Turing Test [1] and started the idea of "Can Machines think?". Basically, the idea is to build a machine which can mimic a human and make another person to believe that he is actually conversing with a human. This popularized the idea of Chatbot.

Then in 1966, the first Chatbot ELIZA [2] was developed by MIT professor Joseph Weizenbaum. ELIZA identified the keywords from user's input and applied a set of decomposition and reassembly rules and generated a response to the user. This Chatbot is based on Pattern Recognition Algorithm and the response was repetitive. In 1995, the Chatbot A.L.I.C.E [3] was developed using pattern matching and AIML – Artificial Intelligence Markup Language. This won the Loebner Prize as "the most human computer" at the annual Turing Test contests in 2000 and 2001. Many considered A.L.I.C.E as an extension of ELIZA. But the major difference is in the categories of Knowledge. A.L.I.C.E had more than 40000 categories of knowledge, whereas the original ELIZA had only 200 categories. In 1990s, a lot of research in this area led to the development of many conversational systems. But they were designed mainly to specialize in a specific domain. The best example is DARPA Airline Travel Information System (ATIS) [4], which was designed to book Airline tickets.

The Chatbot till then were mostly based on pattern matching algorithms or on markup language like AIML. The Chatbot technology took a big stride after the rise of Deep Learning. In this approach, the Chatbot/Neural network was trained on large amount of data, which made it possible for the Bot to generate a reply for user's utterance. Chatbot is a kind of application in which the output of the Model at time t is dependent on all its previous input that it received till time t-1. The basic DNN model does not have any memory to remember its previous input and so it cannot be used for Chatbot.

## 2.2 Deep Learning based Language Models

RNN [5] made its mark with the rise of Deep Learning. RNN model has the ability to take a sequence of input and predict the output based on the inputs it received over a period of time. But the basic implementation of RNN cannot be used as such because it suffered from vanishing and exploding gradient. In 1994, Bengio, et al. [6] clearly described the practical difficulties in training the plain RNN models, when there is a requirement to latch onto the information for long duration. This was overcome by the introduction of LSTM [7] by Hochreiter.

LSTM is a variant of RNN that uses gating mechanism to control the flow of information. All RNN network have a chain of repeating modules. But they had a very basic structure like a single nonlinear layer. In LSTM, they made this layer a little complex by introducing a cell state, forget gate layer and input gate layer. Forget gate controls what information we need to delete from the cell state and input gate layer decides what information need to be added to the cell state. Many other variants of LSTM were introduced after this. Ger, et al [8] introduced a variant which connected the cell state with all gates present in LSTM. Cho, et al. introduced Gated Recurrent Unit or GRU [9]. This variant used update and reset gates and unlike LSTM, this did not have 2 different states (Cell State and Hidden State). In 2015, Yao, et al. introduced Depth – Gated Recurrent Units [10]. This employed a new depth gate to control the flow of information from lower memory cell to upper memory cell. Similarly, many more variants of LSTM were introduced. In 2017, Greff, et al. did a detailed study [11] on the performance of different LSTM variants on three representative tasks: speech recognition, handwriting recognition, and polyphonic music modeling. They summarized the results of 5400 experimental runs and concluded that the plain vanilla LSTM and GRU performs better than all the other variants.

Then seq2seq models are build using Encoder and Decoder technique. In seq2seq method, the input sentence is encoded into fixed length vectors and this was seen as a drawback. So a slightly advanced mechanism was introduced which uses Attention Mechanism [12]. RNN and LSTM process the inputs sequentially word by word. This makes the training of these models as a time consuming task. The most recent Architecture in this field is Transformer Model [13], the first transduction model based entirely on attention, replacing the recurrent layers most commonly used in encoder-decoder architectures with multi-headed self-attention.

## 2.3 Selection of Framework

Instead of building the ChatBot from scratch, we are about to use an existing Framework. In market there are several widely used NLUs which can be easily integrated with third-party applications namely IBM Watson, Dialogflow, Rasa, and Microsoft LUIS. They are popular and widely used by both researchers and practitioners and have been studied by prior NLU comparison work in other domains [14]. Moreover, all selected NLUs can be trained by importing the data through their user interface or API calls, which facilitates the training process.

- Watson Conversation (IBM Watson): An NLU provided by IBM. IBM Watson has prebuilt models for different domains (e.g., healthcare) and a visual dialog editor to simplify building the dialog by non-programmers.
- Dialogflow: An NLU developed by Google. Dialogflow supports more than 20 spoken languages and can be integrated with many chatting platforms such as Slack.
- Rasa: The "only open-source NLU" in our study, owned by Rasa Technologies [15]. Rasa allows developers to configure, deploy, and run the NLU on local servers. Thus, increasing the processing speed by saving the network time compared to cloud-based platforms. In our evaluation, we use Rasa Core v2.0, which was the latest version while exploring.
- Language Understanding Intelligent Service (LUIS): An NLU cloud platform from Microsoft. LUIS has several prebuilt domains such as music and weather, and supports five programming languages: C#, Go, Java, Node.js, and Python.

Based on pervious study [16], all the above platforms perform reasonably well with their tasks. But Rasa is the only Open Source, which allows developers to configure, deploy, and run the NLU on local servers.

### 2.4 Rasa Open Source Architecture

The Rasa Open Source has 2 main components – Rasa NLU and Rasa Core.
RASA NLU is the part which takes care of all NLU processing. This part does the processing like tokenization, featurization, intent identification and entity extraction.
RASA Core is the Dialogue Manager of this framework. Through various policies this controls how the conversation flows for every input entered by the user.
RASA uses DIET (Dual Intent and Entity Transformer) [17] Architecture for identifying the user intent and extracting the entity from User data.

Like a typical Chatbot implementation, NLU and Dialogue Management (DM) are the heart of this Framework. Rasa NLU is just treated like an ear which is taking inputs from user and Rasa Core (Dialogue Management) is just like the brain which will take decisions based on user input.

Each words/tokens of the utterances are handled by two paths
    a.) Pretrained Embedding:
        This can be any of the pretrained embedding [18] like Glove [19] or BERT [20]. This outputs dense features for user messages

b.) Sparse Features and Feed Forward Network:
This output Sparse Features of user messages using one hot encoding and character n-gram methods and these are then passed through a Feed Forward Network.

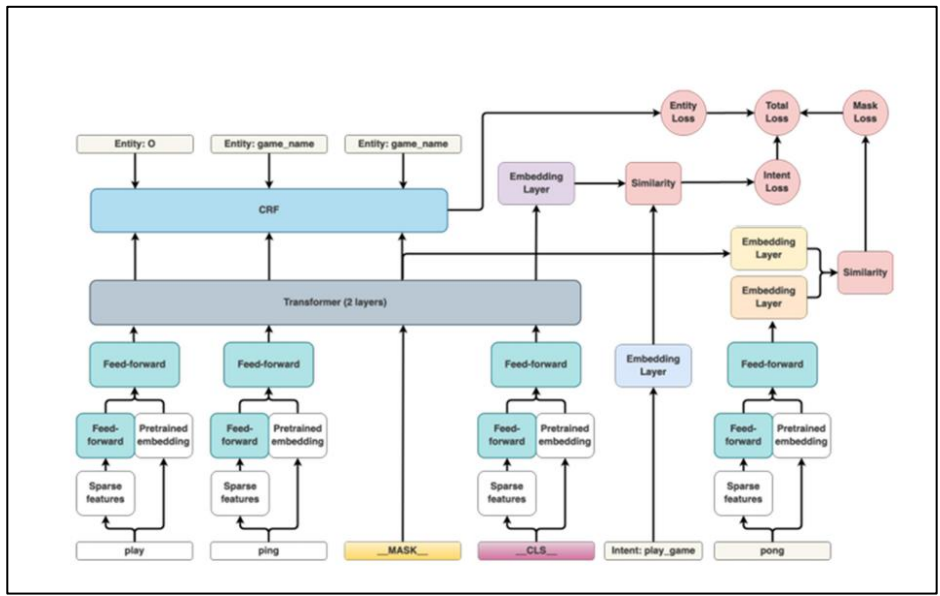

**Fig. 1.** DIET Architectural Diagram

The inputs given to the Feed Forward are sparse in nature and this Architecture is designed to drop nearly 80% of those values.
The DIET has a special CLS token which summarizes the entire utterance or sentence. This CLS token is used to find the intent. In the above architecture diagram, you can see that the CLS token passed through the two paths are checked for similarity with the intent which gets passed through Tensor flow Embedding Layer. This loss is called "Intent Loss".

This architecture also has a MASK token. This is used for generalizing the Model to handle the unseen samples in efficient way. Words from user utterances are masked and then it goes through Transformer [21] and Embedding layer. This value is compared with the actual token which is masked. The similarity calculation at this point has a loss called "Mask Loss".

DIET architecture uses 2 layers Transformer with relative position attention [22] to encode context across the complete sentence. The transformer architecture requires its input to be the same dimension as the transformer layers. Therefore, the concatenated features are passed through another fully connected layer with shared weights across all sequence steps to match the dimension of the transformer layers

The Entity labels are predicted through a Conditional Random Field (CRF) [23] tagging layer on top of Transformer output sequence. The loss calculated at this point is called Entity loss.
The total loss is a combination of all 3 losses – Intent, Mask and Entity Loss. During training the model is optimized to reduce this total loss. The loss is calculated using dot-product [24, 25] methodology.

Rasa Core controls the Dialogue Management of the ChatBot. This is achieved through various policies. The key policy behind this is TED policy. Transformer Embedding Dialogue (TED) [26] is a dialogue policy based on Transformers. TED greatly simplifies the architecture of REDP [27] and do not use a classifier to select system action. Instead, it jointly train embedding for the dialogue state and each of the
System actions by maximizing a similarity function between them

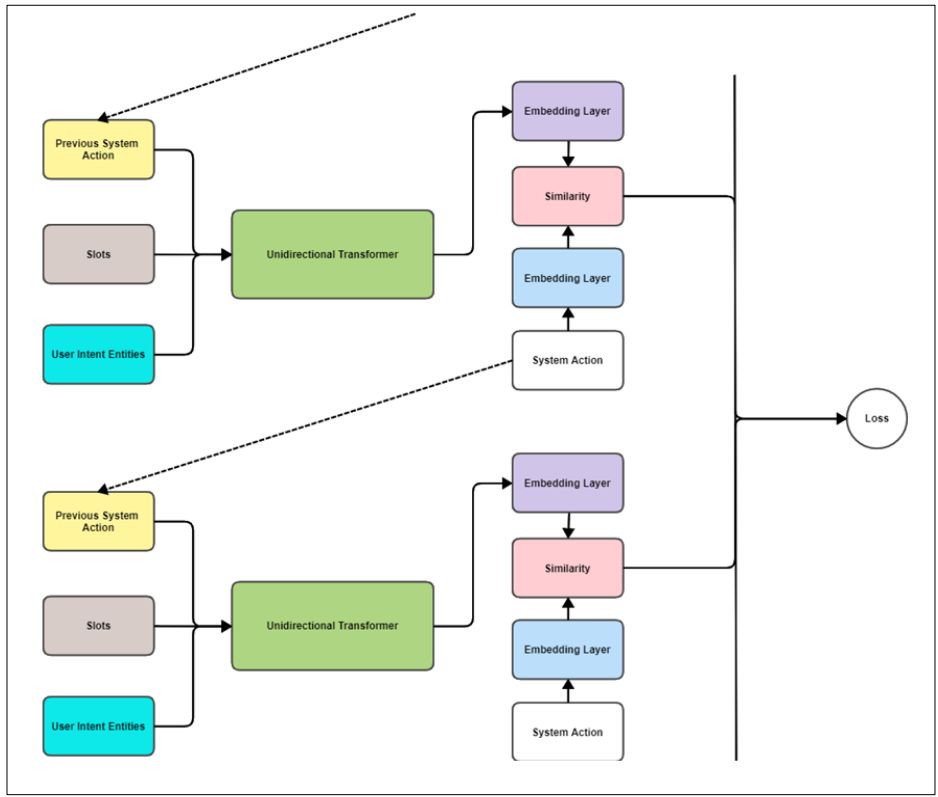

**Fig. 2.** TED Policy Architectural Diagram

## 3. Farmer's Assistant Design

### 3.1 Functional Architecture

Collect the Agricultural data from multiple sources. The data is then processed and segregated into the Training Data and Knowledge Base. The knowledge base is stored in a database like MySQL, MangoDB or SQLITE Database. The training data is used for training the ChatBot to find the Intent and Entity of user utterances. The ChatBot is integrated with an interface and shared to the users.

When the user enters their query in the UI, the ChatBot identifies the intent and entity and retrieves the appropriate response from the Knowledge base. The conversations are tracked and stored in the Database. Various reports created out of this are used to improve the performance of ChatBot.

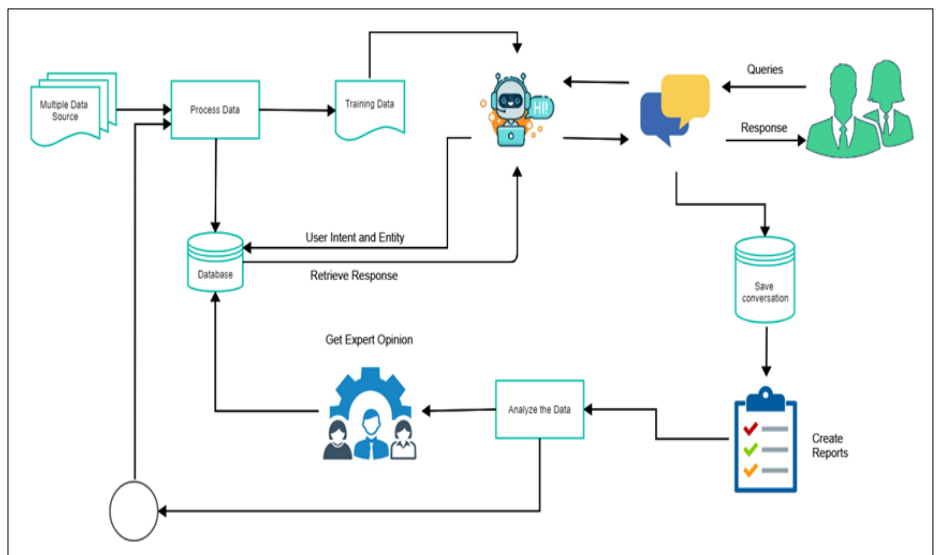

**Fig. 3.** Functional Architecture of proposed system – Farmer's Assistant

### 3.2 Process Flow of ChatBot Model

The user enters the data into ChatBot in Natural Language. This cannot be processed as such by our Model. They need to be tokenized and then converted into numerical vectors. There are techniques like skip-gram, word2vec[28] and CBOW (Continuous Bag of Words) to convert words to vectors. Once the words are converted into vectors, then the next step is intent classification. The intent classification is nothing but understanding the intent or purpose of the user from the Natural Language that she/he inputs. This can be considered as a Machine Learning Classification problem using SVM or we can use RNN-LSTM. The entity extraction is nothing but extraction key information from the user utterance. The Entity recognition can be done using Spacy-NER, or using RNN-LSTM [29] or by using bidirectional LSTM, CNN and CRF.

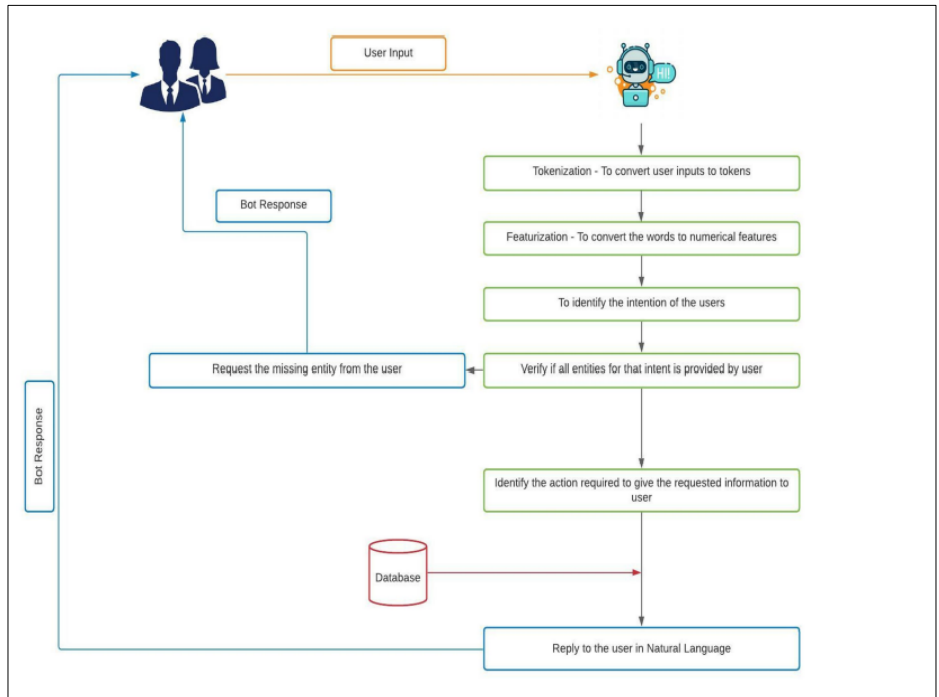

**Fig. 4.** Process Flow

The above discussed steps like tokenization, featurization, Intent classification or Entity extraction is done in Rasa Open Source using the Rasa NLU pipeline. The Dialogue Management is taken care by Rasa Core policies. The various custom actions like retrieving the data from database, interacting with external API to get the data are done through Rasa SDK and Action server.

### 3.3 Data used

We collected the Kisan Call Center data from Open Government Data platform India – data.gov.in. Three years' worth of data from the state Tamil Nadu alone was considered for our analysis. Nearly 51% of the queries are for RABI season crops. These queries are high during the first 3 months of the year and less during the last 4 months of the year. The questions regarding the crop "Paddy" is the major contributor. The queries came from 32 unique districts. The top 15 districts contributed to nearly 75% of all queries. There were queries in 63 different querytypes. But top 15 querytype contribute to nearly 93% of all queries with "Plant Protection" and "Nutrient Management" remaining the top 2. We analyzed the top 15 querytypes in top 6 districts and noticed that "Plant Protection" and "Nutrient Management" remained in top 2 positions in all districts.

These analyses fairly give an idea of how the demand is across the year and how the load should be managed. It also lets us know which crops, querytype and demographic details should be considered for preparing the Knowledge Base.

### 3.4 Knowledge Base

We analyzed the data and prepared the knowledge base with the help of Domain SME - Dr. N. Murugesan [Retd Professor - TNAU]. The data are stored in SQLITE database for plant protection, nutrient management and Agricultural Officers contact.

**Table 1.** Plant Protection Knowledge Base.

| Field | Description | Data Type |
| --- | --- | --- |
| *crop* | *Name of the crop – Obtained from User Entity* | *CHAR(30)* |
| *disease* | *Name of the disease – Obtained from User Entity* | *CHAR(60)* |
| *Remedy/Fix* | *Remedy for the disease. Populated based on past data and SME advice* | *VARCHAR(255)* |

**Table 2.** Nutrient Management Knowledge Base.

| Field | Description | Data Type |
| --- | --- | --- |
| *crop* | *Name of the crop – Obtained from User Entity* | *CHAR(30)* |
| *nutrient* | *Name of the nutrient – Obtained from User Entity* | *CHAR(60)* |
| *Remedy/Fix* | *Remedy for the nutrient deficiency. Populated based on past data and SME advice* | *VARCHAR(255)* |

**Table 3.** Agricultural Officers contact Knowledge Base.

| Field | Description | Data Type |
| --- | --- | --- |
| *role* | *Designation of Officers – Obtained from User Entity* | *CHAR(50)* |
| *city* | *City/District they are in charge of – Obtained from User Entity* | CHAR(30) |
| *Phone* | *Their Official Contact number – Populated from Government Websites and Database* | CHAR(38) |
| *mail* | *Official Mail ID for the officers* | CHAR(50) |

## 4. Farmer's Assistant Evaluation

We split the data into training and testing and compared the performance with 4 different configurations files.

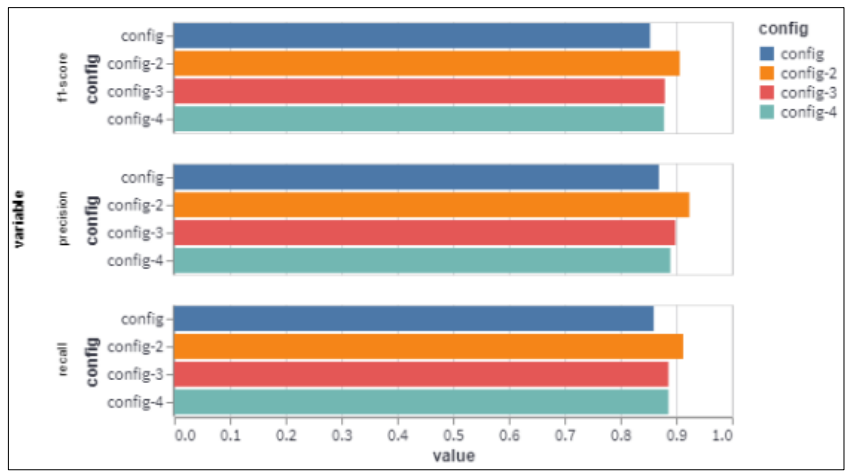

**Fig. 5.** Test Configurations Config, Config-1, Config-2 & Config-3 (ordered from left to right)

**Fig. 6.** Precision, recall and f1-score comparison for Intent Identification of all 4 configurations

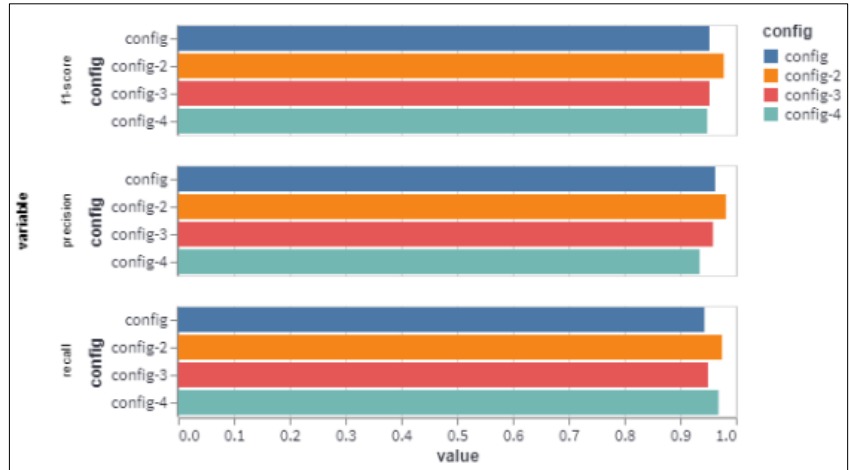

**Fig. 7.** Precision, recall and f1-score comparison for Entity Identification of all 4 configurations

From the above graph, it is clear that the model which uses both sparse and pre-embedding technique in featurization outperforms the other models.

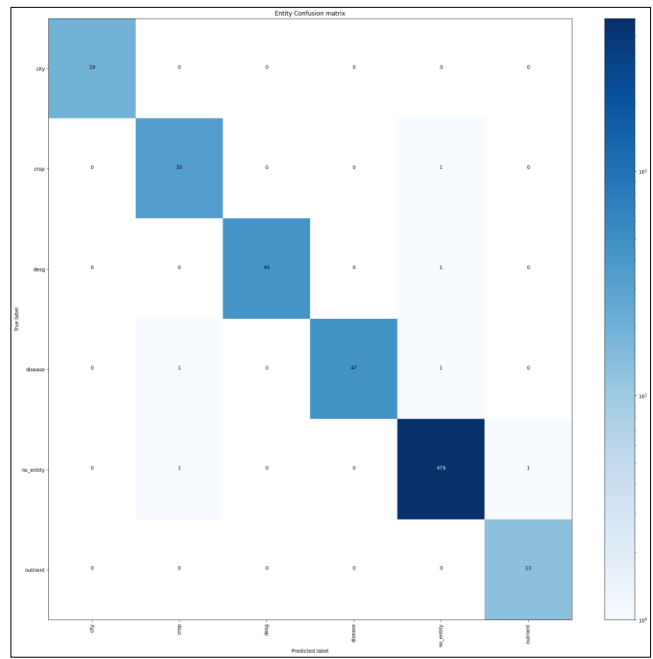

**Fig. 8.** Entitiy confusion Matrix

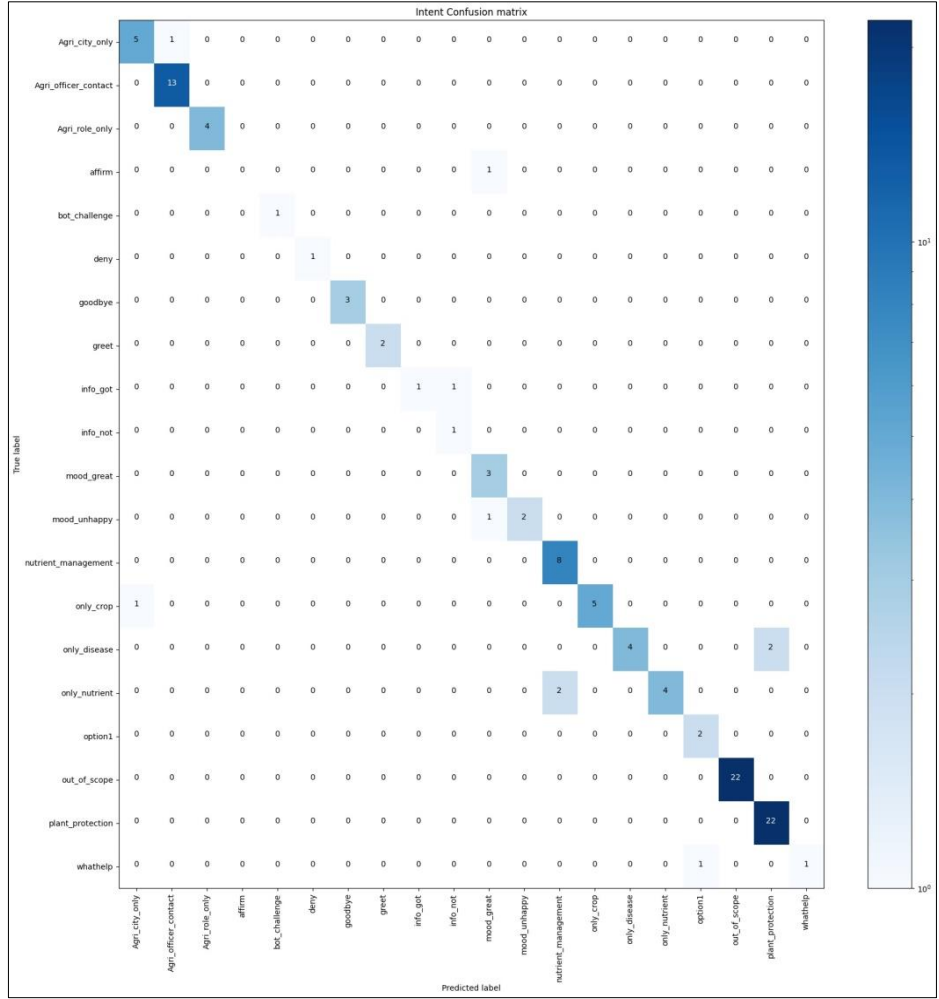

**Fig. 9.** Intent confusion Matrix

The intent and entity confusion matrix helped us to understand the intent and entity, which couldn't be found by our Model.

## 5. Result

Farmer's Assistant was able to successfully converse with the user and identify the intent and entity for which it was trained. The Bot retrieved appropriate response from the database and shared it with the user. It worked well even for questions which are out of scope for the Bot's functionality.

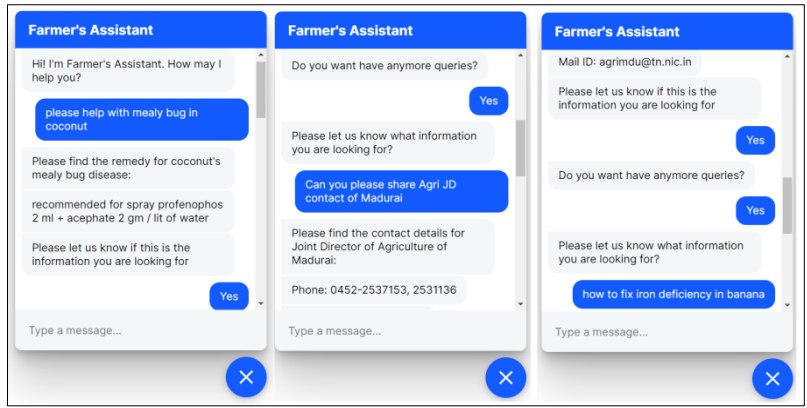

**Fig. 10.** Snapshots of Farmer's Assistant when the information is available

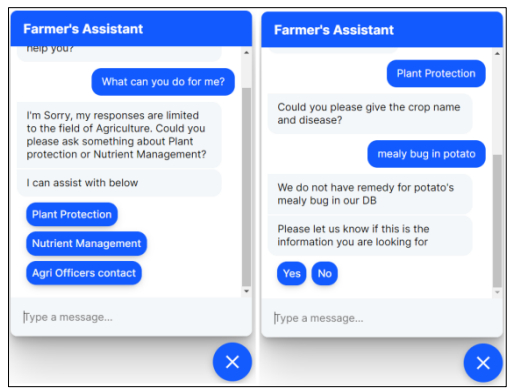

**Fig. 11.** Snapshots of Farmer's Assistant when the date is not available

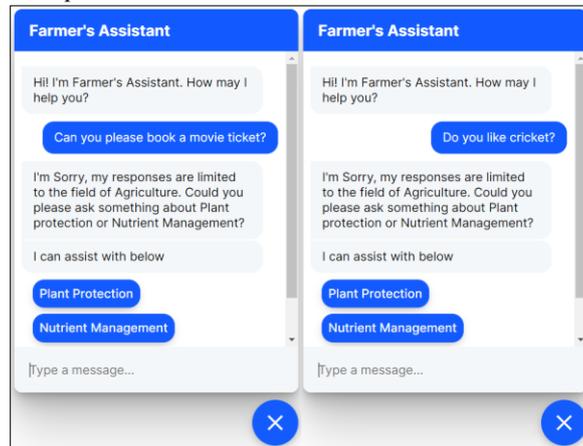

**Fig. 12.** Snapshots of Farmer's Assistant for out of scope queries

## 6. **Limitations**

This model supports only English. But we cannot expect all the users to know English. The lack of multilingual support is a serious limitation of this initiative. Most of the times, the users will not follow just the usual conversation flow. The possibility of getting return questions, for the recommendations that we give from our repository, is very high. We cannot plan all these scenarios in advance and in such cases the Bot can only handoff the conversation to a human.

The limitations can be overcome by including multilingual support in the future. A reporting mechanism needs to be created to find the conversation which did not give expected answers to the user. Based on this report, the data for the ChatBot used should be updated to prevent this scenario from happening again.

## 7. **Conclusion**

The lack of consistent medium to pass on the experts knowledge to Farmers is the open problem in Agriculture based country like India. We were able to use the data from KCC Dataset and Domain Expert knowledge and build a ChatBot to answer the basic queries of Farmer's in an unbiased manner. The performance metrics of the ChatBot also showed promising result for the data in which it was trained. To implement this in large scale, we need a committee of experts to collect the data to strengthen our Knowledge base. Moreover addition of image recognition feature like identifying plant disease based on photograph and suggesting remedy for the same will add high values to this initiative.